\documentclass[runningheads]{llncs}

\usepackage{tikz}
\usetikzlibrary{arrows,shapes,calc,matrix,fit,backgrounds}
\usepackage{pgfplots}
\usepackage{pgfplotstable}
\pgfplotsset{compat=1.9}

\usepackage{xstring}

\usepgfplotslibrary{external}
\newcommand{\extdata}[1]{\input{#1}}
\IfBeginWith*{\jobname}{fig/extern/}{\finalcopy}{}

\newcommand{\leg}[1]{\addlegendentry{#1}}

\tikzset{every mark/.append style={solid}}
\pgfplotsset{%smooth,
	grid=both, width=\columnwidth, try min ticks=5,
	every axis/.append style={font=\scriptsize},
	every axis plot/.append style={thick,mark=none,mark size=1.2,tension=0.18},
	legend cell align=left, legend style={fill opacity=0.8},
}

\pgfplotsset{
	dash/.style={mark=o,dashed,opacity=0.7},
	dott/.style={mark=o,dotted,opacity=0.7},
}

\usepackage{graphicx}
\usepackage{comment}
\usepackage{amsmath,amssymb} % define this before the line numbering.

\usepackage{color}
\usepackage{epsfig}
\usepackage{graphicx}
\usepackage[normalem]{ulem}
\usepackage{algorithm}
\usepackage[noend]{algpseudocode}
\usepackage{booktabs}

\usepackage{pifont}
\usepackage{multirow}
\usepackage{xspace}

\usepackage{pgfkeys}

\usepackage[symbol]{footmisc} % Change footnotes to symbols
\begin{document}
\pagestyle{headings}
\mainmatter
\def\ECCVSubNumber{2614}  % Insert your submission number here

\title{Memory-Efficient Incremental Learning Through Feature Adaptation} % Replace with your title

%\begin{comment}
\titlerunning{Memory-Efficient Incremental Learning Through Feature Adaptation}
\author{Ahmet Iscen\inst{1}\and
Jeffrey Zhang\inst{2}\and
Svetlana Lazebnik\inst{2}\and
Cordelia Schmid\inst{1}
}
\authorrunning{A. Iscen et al.}
\institute{Google Research \and
University of Illinois at Urbana-Champaign
}
%\end{comment}
%******************
\maketitle

\newcommand{\nn}[1]{\ensuremath{\text{NN}_{#1}}\xspace}

\newcommand{\citemiss}{\alert{[??]}\xspace}

\newcommand{\supe}[1]{^{\mkern-2mu(#1)}}
\newcommand{\dime}[1]{(#1)}

\def\l1{\ensuremath{\ell_1}\xspace}
\def\l2{\ensuremath{\ell_2}\xspace}

\newcommand*\OK{\ding{51}}

\newenvironment{narrow}[1][1pt]
	{\setlength{\tabcolsep}{#1}}
	{\setlength{\tabcolsep}{6pt}}

%--------------------------------------------------------------------------------
%algorithm
\newcommand{\algocom} [1]{{\color{orange} \Comment     #1}} % colored comment
\newcommand{\commentout}[1]{}
\newcommand{\prm}[1]{_{#1}}

%--------------------------------------------------------------------
\newcommand{\alert}[1]{{\color{red}{#1}}}
\newcommand{\head}[1]{{\smallskip\noindent\bf #1}}
\newcommand{\equ}[1]{(\ref{equ:#1})\xspace}

\newcommand{\red}[1]{{\color{red}{#1}}}
\newcommand{\blue}[1]{{\color{blue}{#1}}}
\newcommand{\green}[1]{{\color{green}{#1}}}
\newcommand{\gray}[1]{{\color{gray}{#1}}}

%--------------------------------------------------------------------

\newcommand{\tran}{^\top}
\newcommand{\mtran}{^{-\top}}
\newcommand{\zcol}{\mathbf{0}}
\newcommand{\zrow}{\zcol\tran}

\newcommand{\ind}{\mathbbm{1}}
\newcommand{\expect}{\mathbb{E}}
\newcommand{\nat}{\mathbb{N}}
\newcommand{\zahl}{\mathbb{Z}}
\newcommand{\real}{\mathbb{R}}
\newcommand{\proj}{\mathbb{P}}
\newcommand{\prob}{\mathbf{Pr}}

\newcommand{\mif}{\textrm{if }}
\newcommand{\other}{\textrm{otherwise}}
\newcommand{\minimize}{\textrm{minimize }}
\newcommand{\maximize}{\textrm{maximize }}

\newcommand{\id}{\operatorname{id}}
\newcommand{\const}{\operatorname{const}}
\newcommand{\sgn}{\operatorname{sgn}}
\newcommand{\var}{\operatorname{Var}}
\newcommand{\mean}{\operatorname{mean}}
\newcommand{\trace}{\operatorname{tr}}
\newcommand{\diag}{\operatorname{diag}}
\newcommand{\vect}{\operatorname{vec}}
\newcommand{\cov}{\operatorname{cov}}

\newcommand{\softmax}{\operatorname{softmax}}
\newcommand{\clip}{\operatorname{clip}}

\newcommand{\defn}{\mathrel{:=}}
\newcommand{\peq}{\mathrel{+\!=}}
\newcommand{\meq}{\mathrel{-\!=}}

\newcommand{\floor}[1]{\left\lfloor{#1}\right\rfloor}
\newcommand{\ceil}[1]{\left\lceil{#1}\right\rceil}
\newcommand{\inner}[1]{\left\langle{#1}\right\rangle}
\newcommand{\norm}[1]{\left\|{#1}\right\|}
\newcommand{\frob}[1]{\norm{#1}_F}
\newcommand{\card}[1]{\left|{#1}\right|\xspace}
\newcommand{\diff}{\mathrm{d}}
\newcommand{\der}[3][]{\frac{d^{#1}#2}{d#3^{#1}}}
\newcommand{\pder}[3][]{\frac{\partial^{#1}{#2}}{\partial{#3^{#1}}}}
\newcommand{\ipder}[3][]{\partial^{#1}{#2}/\partial{#3^{#1}}}
\newcommand{\dder}[3]{\frac{\partial^2{#1}}{\partial{#2}\partial{#3}}}

\newcommand{\wb}[1]{\overline{#1}}
\newcommand{\wt}[1]{\widetilde{#1}}

\def\xxssp{\hspace{-3pt}}
\def\xssp{\hspace{1pt}}
\def\ssp{\hspace{3pt}}
\def\msp{\hspace{5pt}}
\def\lsp{\hspace{12pt}}

\newcommand{\cA}{\mathcal{A}}
\newcommand{\cB}{\mathcal{B}}
\newcommand{\cC}{\mathcal{C}}
\newcommand{\cD}{\mathcal{D}}
\newcommand{\cE}{\mathcal{E}}
\newcommand{\cF}{\mathcal{F}}
\newcommand{\cG}{\mathcal{G}}
\newcommand{\cH}{\mathcal{H}}
\newcommand{\cI}{\mathcal{I}}
\newcommand{\cJ}{\mathcal{J}}
\newcommand{\cK}{\mathcal{K}}
\newcommand{\cL}{\mathcal{L}}
\newcommand{\cM}{\mathcal{M}}
\newcommand{\cN}{\mathcal{N}}
\newcommand{\cO}{\mathcal{O}}
\newcommand{\cP}{\mathcal{P}}
\newcommand{\cQ}{\mathcal{Q}}
\newcommand{\cR}{\mathcal{R}}
\newcommand{\cS}{\mathcal{S}}
\newcommand{\cT}{\mathcal{T}}
\newcommand{\cU}{\mathcal{U}}
\newcommand{\cV}{\mathcal{V}}
\newcommand{\cW}{\mathcal{W}}
\newcommand{\cX}{\mathcal{X}}
\newcommand{\cY}{\mathcal{Y}}
\newcommand{\cZ}{\mathcal{Z}}

\newcommand{\vA}{\mathbf{A}}
\newcommand{\vB}{\mathbf{B}}
\newcommand{\vC}{\mathbf{C}}
\newcommand{\vD}{\mathbf{D}}
\newcommand{\vE}{\mathbf{E}}
\newcommand{\vF}{\mathbf{F}}
\newcommand{\vG}{\mathbf{G}}
\newcommand{\vH}{\mathbf{H}}
\newcommand{\vI}{\mathbf{I}}
\newcommand{\vJ}{\mathbf{J}}
\newcommand{\vK}{\mathbf{K}}
\newcommand{\vL}{\mathbf{L}}
\newcommand{\vM}{\mathbf{M}}
\newcommand{\vN}{\mathbf{N}}
\newcommand{\vO}{\mathbf{O}}
\newcommand{\vP}{\mathbf{P}}
\newcommand{\vQ}{\mathbf{Q}}
\newcommand{\vR}{\mathbf{R}}
\newcommand{\vS}{\mathbf{S}}
\newcommand{\vT}{\mathbf{T}}
\newcommand{\vU}{\mathbf{U}}
\newcommand{\vV}{\mathbf{V}}
\newcommand{\vW}{\mathbf{W}}
\newcommand{\vX}{\mathbf{X}}
\newcommand{\vY}{\mathbf{Y}}
\newcommand{\vZ}{\mathbf{Z}}

\newcommand{\va}{\mathbf{a}}
\newcommand{\vb}{\mathbf{b}}
\newcommand{\vc}{\mathbf{c}}
\newcommand{\vd}{\mathbf{d}}
\newcommand{\ve}{\mathbf{e}}
\newcommand{\vf}{\mathbf{f}}
\newcommand{\vg}{\mathbf{g}}
\newcommand{\vh}{\mathbf{h}}
\newcommand{\vi}{\mathbf{i}}
\newcommand{\vj}{\mathbf{j}}
\newcommand{\vk}{\mathbf{k}}
\newcommand{\vl}{\mathbf{l}}
\newcommand{\vm}{\mathbf{m}}
\newcommand{\vn}{\mathbf{n}}
\newcommand{\vo}{\mathbf{o}}
\newcommand{\vp}{\mathbf{p}}
\newcommand{\vq}{\mathbf{q}}
\newcommand{\vr}{\mathbf{r}}
\newcommand{\vt}{\mathbf{t}}
\newcommand{\vu}{\mathbf{u}}
\newcommand{\vv}{\mathbf{v}}
\newcommand{\vw}{\mathbf{w}}
\newcommand{\vx}{\mathbf{x}}
\newcommand{\vy}{\mathbf{y}}
\newcommand{\vz}{\mathbf{z}}

\newcommand{\vone}{\mathbf{1}}
\newcommand{\vzero}{\mathbf{0}}

\newcommand{\valpha}{{\boldsymbol{\alpha}}}
\newcommand{\vbeta}{{\boldsymbol{\beta}}}
\newcommand{\vgamma}{{\boldsymbol{\gamma}}}
\newcommand{\vdelta}{{\boldsymbol{\delta}}}
\newcommand{\vepsilon}{{\boldsymbol{\epsilon}}}
\newcommand{\vzeta}{{\boldsymbol{\zeta}}}
\newcommand{\veta}{{\boldsymbol{\eta}}}
\newcommand{\vtheta}{{\boldsymbol{\theta}}}
\newcommand{\viota}{{\boldsymbol{\iota}}}
\newcommand{\vkappa}{{\boldsymbol{\kappa}}}
\newcommand{\vlambda}{{\boldsymbol{\lambda}}}
\newcommand{\vmu}{{\boldsymbol{\mu}}}
\newcommand{\vnu}{{\boldsymbol{\nu}}}
\newcommand{\vxi}{{\boldsymbol{\xi}}}
\newcommand{\vomikron}{{\boldsymbol{\omikron}}}
\newcommand{\vpi}{{\boldsymbol{\pi}}}
\newcommand{\vrho}{{\boldsymbol{\rho}}}
\newcommand{\vsigma}{{\boldsymbol{\sigma}}}
\newcommand{\vtau}{{\boldsymbol{\tau}}}
\newcommand{\vupsilon}{{\boldsymbol{\upsilon}}}
\newcommand{\vphi}{{\boldsymbol{\phi}}}
\newcommand{\vchi}{{\boldsymbol{\chi}}}
\newcommand{\vpsi}{{\boldsymbol{\psi}}}
\newcommand{\vomega}{{\boldsymbol{\omega}}}

\newcommand{\rLambda}{\mathrm{\Lambda}}
\newcommand{\rSigma}{\mathrm{\Sigma}}

%--------------------------------------------------------------------
% Add a period to the end of an abbreviation unless there's one
% already, then \xspace.
% \makeatletter
% \DeclareRobustCommand\onedot{\futurelet\@let@token\@onedot}
% \def\@onedot{\ifx\@let@token.\else.\null\fi\xspace}
\def\onedot{.\xspace}
\def\eg{\emph{e.g}\onedot} \def\Eg{\emph{E.g}\onedot}
\def\ie{\emph{i.e}\onedot} \def\Ie{\emph{I.e}\onedot}
\def\cf{\emph{cf}\onedot} \def\Cf{\emph{C.f}\onedot}
\def\etc{\emph{etc}\onedot}
\def\vs{\emph{vs}\onedot}
\def\wrt{w.r.t\onedot} \def\dof{d.o.f\onedot}
\def\etal{\emph{et al}.}

\newcommand{\std}[1]{\tiny{$\pm$#1}}

\makeatother

\begin{abstract}
We introduce an approach for incremental learning that preserves feature descriptors of training images from previously learned classes, instead of the images themselves, unlike most existing work. Keeping the much lower-dimensional feature embeddings of images reduces the memory footprint significantly. 
We assume that the model is updated incrementally for new classes as new data becomes available sequentially.
This requires adapting the previously stored feature vectors to the updated feature space without having access to the corresponding original training images.
Feature adaptation is learned with a multi-layer perceptron, which is trained on feature pairs corresponding to the outputs of the original and updated network on a training image.
We validate experimentally that such a transformation generalizes well to the features of the previous set of classes, and maps features to a discriminative subspace in the feature space.
As a result, the classifier is optimized jointly over new and old classes without requiring old class images. 
Experimental results show that our method achieves state-of-the-art classification accuracy in incremental learning benchmarks, while having at least an order of magnitude lower memory footprint compared to image-preserving strategies.
\end{abstract}

\section{Introduction}

Deep neural networks have shown excellent performance for many computer vision problems, such as image classification~\cite{HZRS16,KSH12,SZ14} and object detection~\cite{HGD+17,RHG15}. 
However, most common models require large amounts of labeled data for training, and assume that data from all possible classes is available for training at the same time. 

By contrast, \emph{class incremental learning}~\cite{RKS+17} addresses the setting where training data is received sequentially, and data from previous classes is discarded as data for new classes becomes available. 
Thus, classes are not learned all at once.
Ideally, models should learn the knowledge from new classes while maintaining the knowledge learned from previous classes.
This poses a significant problem, as neural networks are known to quickly forget what is learned in the past -- a phenomenon known as \emph{catastrophic forgetting}~\cite{MC89}.
Recent approaches alleviate catastrophic forgetting in neural networks by adding regularization terms that encourage the network to stay similar to its previous states~\cite{KPR+17,LH17} or by preserving a subset of previously seen data~\cite{RKS+17}.

\begin{figure*}[t]
\centering
\includegraphics[width=0.90\textwidth]{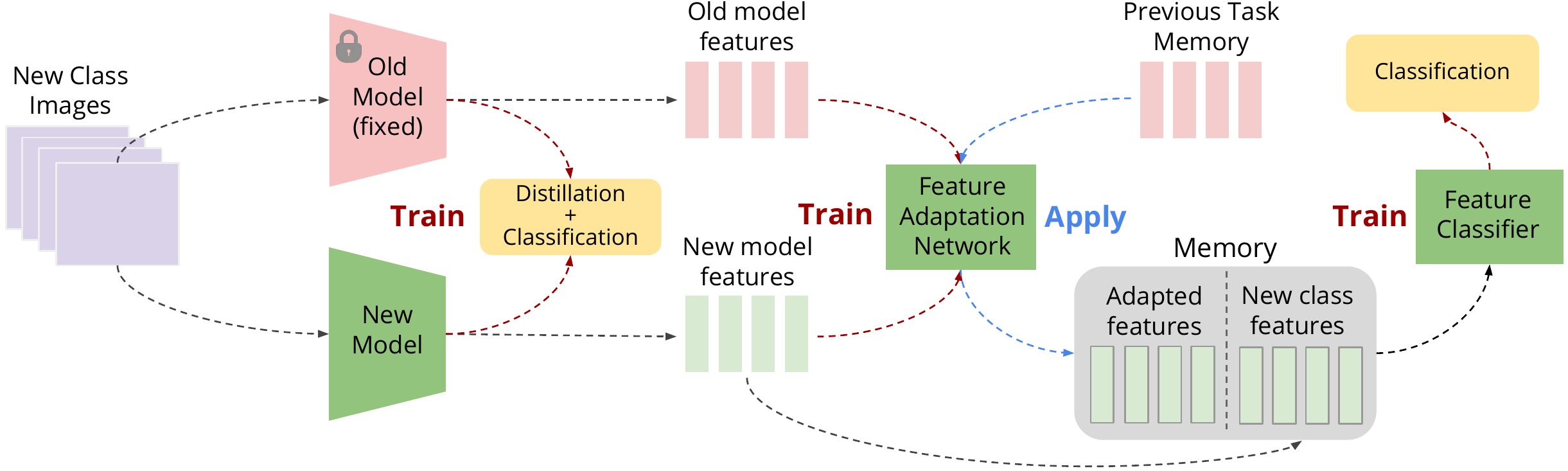}
\caption{An overview of our method. Given new class images, a new model is trained on the data with distillation and classification losses. Features are extracted using the old and new models from new class images to train a feature adaptation network. The learned feature adaptation network is applied to the preserved vectors to transform them into the new feature space. With features from all seen classes represented in the same feature space, we train a feature classifier.
\label{fig:overview}
}
\end{figure*}
One of the criteria stated by Rebuffi~\etal~\cite{RKS+17} for a successful incremental learner is that ``computational requirements and memory footprint should remain bounded, or at least grow very slowly, with respect to the number of classes seen so far''.
In our work, we significantly improve the memory footprint required by an incremental learning system.
We propose to preserve a subset of \emph{feature descriptors} rather than images. 
This enables us to compress information from previous classes in low-dimensional embeddings. 
For example, for ImageNet classification using ResNet-18, storing a $512$-dimensional feature vector has $\sim1\%$ of the storage requirement compared to storing a $256\times256\times3$ image (Sec. \ref{sec:sota}).
Our experiments show that we achieve better classification accuracy compared to state-of-the-art methods, with a memory footprint of at least an order of magnitude less.

Our strategy of preserving feature descriptors instead of images faces a serious potential problem:
as the model is trained with more classes, the feature extractor changes, making the preserved feature descriptors from previous feature extractors obsolete. To overcome this difficulty, we propose a \emph{feature adaptation} method that learns a mapping between two feature spaces. 
As shown in Fig.\ref{fig:overview}, our novel approach allows us to learn the changes in the feature space and adapt the preserved feature descriptors to the new feature space. 
With all image features in the same feature space,  we can train a feature classifier to correctly classify features from all seen classes.
To summarize, our contributions in this paper are as follows:
\begin{itemize}
\item We propose an incremental learning framework where previous feature descriptors, instead of previous images, are preserved.
\item We present a \emph{feature adaptation} approach which maps previous feature descriptors to their correct values as the model is updated.
\item We apply our method on popular class-incremental learning benchmarks and show that we achieve top accuracy on ImageNet compared to other state-of-the-art methods while significantly reducing the memory footprint.
\end{itemize}
\section{Related work}
The literature for \emph{incremental learning} prior to the deep-learning era includes incrementally trained support vector machines~\cite{CP01}, random forests~\cite{RGG+15}, and metric-based methods that generalize to new classes~\cite{MVP+13}.
We restrict our attention mostly to more recent deep-learning-based methods.
Central to most of these methods is the concept of {\em rehearsal}, which is defined as preserving and replaying data from previous sets of classes when updating the model with new classes~\cite{R95}. 

\head{Non-rehearsal methods } 
do not preserve any data from previously seen classes. 
Common approaches include increasing the network capacity for new sets of classes~\cite{RRD+16,WRH+17}, or weight consolidation, which identifies the important weights for previous sets of classes and slows down their learning~\cite{KPR+17}. 
Chaudhry~\etal~\cite{CDA+18} improve weight consolidation by adding KL-divergence-based regularization. 
Liu~\etal~\cite{LMH+18} rotate the parameter space of the network and show that the weight consolidation is more effective in the rotated parameter space.
Aljundi~\etal~\cite{ABE+18} compute the importance of each parameter in an unsupervised manner without labeled data.
Learning without Forgetting (LwF)~\cite{LH17} (discussed in more detail in Sec.~\ref{sec:background}) reduces catastrophic forgetting by adding a knowledge distillation~\cite{HVD15} term in the loss function, which encourages the network output for new classes to be close to the original network output.
Learning without Memorizing~\cite{DSP+19} extends LwF by adding a distillation term based on attention maps.
Zhang~\etal~\cite{ZZG+19} argue that LwF produces models that are either biased towards old or new classes.
They train a separate model for new classes, and consolidate the two models with unlabeled auxiliary data. Lastly, Yu~\etal~\cite{YTL+20} updates previous class centroids for NME classification \cite{RKS+17} by estimating the feature representation shift using new class centroids.

\head{Rehearsal with exemplars.}
Lopez-Paz and Ranzato~\cite{LR17} add constraints on the gradient update, and transfer information to previous sets of classes while learning new sets of classes.
Incremental Classifier and Representation Learning (iCARL) by Rebuffi \etal~\cite{RKS+17} preserves a subset of images, called \emph{exemplars}, and includes the selected subset when updating the network for new sets of classes.
Exemplar selection is done with an efficient algorithm called herding~\cite{W09}.
The authors also show that the classification accuracy increases when the mean class vector~\cite{MVP+13} is used for classification instead of the learned classifier of the network.
iCARL is one of the most effective existing methods in the literature, and will be considered as our main baseline.
Castro~\etal~\cite{CMM+18} extend iCARL by learning the network and classifier with an end-to-end approach.
Similarly, Javed and Shafait~\cite{JS18} learn an end-to-end classifier by proposing a dynamic threshold moving algorithm.
Other recent work extend iCARL by correcting the bias and introducing additional constraints in the loss function~\cite{BP19,HPL+19,WCW+19}. 

\head{Rehearsal with generated images.} 
These methods use generative models (GANs~\cite{GPM+14}) to generate \emph{fake} images that mimic the past data, and use the generated images when learning the network for new classes~\cite{SLK+17,WHL+18}. 
He~\etal~\cite{HWS+18} use multiple generators to increase capacity as new sets of classes become available.
A major drawback of these methods is that they are either applied to less complex datasets with low-resolution images, or their success depends on combining the generated images with real images.

\head{Feature-based methods.}
Earlier work on \emph{feature generation}, rather than image generation, focuses on zero-shot learning~\cite{BHJ17,XLS+18}.
Kemker~\etal~\cite{KK18} use a dual-memory system which consists of fast-learning memory for new classes and long-term storage for old classes. Statistics of feature vectors, such as the mean vector and covariance matrix for a set of vectors, are stored in the memory.
Xiang~\etal~\cite{XFJ+19} also store feature vector statistics, and learn a feature generator to generate vectors from old classes.
The drawback of these methods~\cite{KK18,XFJ+19} is that they depend on a pre-trained network.
This is different than other methods (LwF, iCARL) where the network is learned from scratch.

In this paper, we propose a method which performs \emph{rehearsal} with \emph{features}.
Unlike existing feature-based methods, we do not \emph{generate} feature descriptors from class statistics. 
We preserve and adapt feature descriptors to new feature spaces as the network is trained incrementally. 
This allows training the network from scratch and does not depend on a pre-trained model as in~\cite{KK18,XFJ+19}.
Compared to existing rehearsal methods, our method has a significantly lower memory footprint by preserving features instead of images.

Our feature adaptation method is inspired by the feature hallucinator proposed by Hariharan and Girschick~\cite{HG17}.
Their method learns intra-class feature transformations as a way of data augmentation in few-shot learning problem.
Our method is quite different as we learn the transformations between feature pairs of the same image, extracted at two different increments of the network.
Finally, whereas Yu~\etal~\cite{YTL+20} uses interpolation to estimate changes for class centroids of features, our feature adaptation method learns a generalizable transformation function for all stored features.
\section{Background on incremental learning}
\label{sec:background}

This section introduces the incremental learning task and summarizes popular strategies for training the network and handling \emph{catastrophic forgetting}, namely, distillation and preservation of old data.

\head{Problem formulation.}
We are given a set of images $\cX$ with labels $\cY$ belonging to classes in $\cC$. This defines the dataset $\cD = \{(x,y)|x \in \cX, y\in\cY \}$.
In class-incremental learning, we want to expand an existing model to classify new classes. 
Given $T$ tasks, we 
split $\cC$ into $T$ subsets $\cC^1, \cC^2, \dots, \cC^T$, where $\cC = \cC^1 \cup \cC^2 \cup \dots \cup \cC^T$ and $\cC^i \cap \cC^j = \emptyset$ for $i \neq j$. 
We define task $t$ as introducing new classes $\cC^t$ using dataset $\cD^t = \{(x,y)|y\in\cC^t\}$.
We denote $\cX^t = \{x| (x,y)\in \cD^t\}$ and $\cY^t = \{y| (x,y)\in \cD^t\}$ as the training images and labels used at task $t$.
The goal is to train a classifier which accurately classifies examples belonging to the new set of classes $\cC^t$, while still being able to correctly classify examples belonging to classes $\cC^i$, where $i < t$.

\head{The classifier.}
The learned classifier is typically a \emph{convolutional neural network} (CNN) denoted by $f\prm{\theta,W}: \cX \rightarrow \real^K$, where $K$ is the number of classes.  
Learnable parameters $\theta$ and $W$ correspond to two components of the network, the \emph{feature extractor} $h\prm{\theta}$ and the \emph{network classifier} $g\prm{W}$.
The feature extractor $h\prm{\theta}: \cX \rightarrow \real^d$ maps an image to a $d$-dimensional feature vector. 
The network classifier $g\prm{W}: \real^d \rightarrow \real^K$ is applied to the output of the feature extractor $h\prm{\theta}$ and outputs a $K$-dimensional vector for each class classification score.
The network $f\prm{\theta,W}$ is the mapping from the input space directly to confidence scores, where $x \in \cX$:  
\begin{equation}
f\prm{\theta,W}(x) \defn g\prm{W}(h\prm{\theta}(x)).
\label{equ:netOutput}
\end{equation}

Training the parameters $\theta$ and $W$ of the network is typically achieved through a loss function, such as cross-entropy loss,
\begin{equation}
L_{\text{CE}}(x,y) \defn - \sum^{K}_{k=1}  y_k \; \text{log}(\vsigma(f\prm{\theta,W}(x))_k),
\label{eq:lce}
\end{equation}
where $y \in \real^K$ is the label vector and $\vsigma$ is either a \emph{softmax} or \emph{sigmoid} function.

In incremental learning, the number of classes our models output increases at each task. $K^t = \sum_i^t |\cC^i|$ denotes the total number of classes at task $t$.
Notice at task $t$, our model is expected to classify $|\cC^t|$ more classes than task $t-1$.
The network $f\prm{\theta,W}^t$ is only trained with $\cX^t$, the data available in the current task.
Nevertheless, the network is still expected to accurately classify any images belonging to the classes from the previous tasks.

\head{Distillation.} 
One of the main challenges in incremental learning is  \emph{catastrophic forgetting}~\cite{GMX+13,MC89}. 
At a given task $t$, we want to expand a previous model's capability to classify new classes $\cC^t$. We train a new model $f\prm{\theta,W}^{t}$ initialized from $f\prm{\theta,W}^{t-1}$.
Before the training of the task, we freeze a copy of $f\prm{\theta,W}^{t-1}$ to use as reference.
We only have access to $\cX^t$ and not to previously seen data $\cX^i$, where $i < t$.
As the network is updated with $\cX^t$ in Eq.~\eqref{eq:lce}, its knowledge of previous tasks quickly disappears due to catastrophic forgetting.
\emph{Learning without Forgetting} (LwF)~\cite{LH17} alleviates this problem by introducing a \emph{knowledge distillation loss}~\cite{HVD15}.
This loss is a modified cross-entropy loss, which encourages the network $f\prm{\theta,W}^{t}$ to mimic the output of the previous task model $f\prm{\theta,W}^{t-1}$:
\begin{equation}
L_{\text{KD}}(x) \defn - \sum^{K^{t-1}}_{k=1} \vsigma(f\prm{\theta,W}^{t-1}(x))_k \; \text{log}( \vsigma(f\prm{\theta,W}^{t}(x))_k),
\label{eq:ldiss}
\end{equation}
where $x \in \cX^t$. 
$L_{\text{KD}}(x)$ encourages the network to make similar predictions to the previous model.
The knowledge distillation loss term is added to the classification loss~\eqref{eq:lce}, resulting in the overall loss function:
\begin{equation}
L(x,y) \defn L_{\text{CE}}(x,y) + \lambda L_{\text{KD}}(x),
\label{eq:lcedist}
\end{equation}
where $\lambda$ is typically set to 1~\cite{RKS+17}.
Note that the network $f\prm{\theta,W}^t$ is continuously updated at task $t$, whereas the network $f\prm{\theta,W}^{t-1}$ remains frozen and will not be stored after the completion of task $t$.

\head{Preserving data of the old classes.}
A common approach is to preserve some images for the old classes and use them when training new tasks~\cite{RKS+17}. 
At task $t$, new class data refers to $\cX^t$ and old class data refers to data seen in previous tasks, \ie $\cX^i$ where $i<t$.
After each task $t$, a new \emph{exemplar} set $\cP^t$ is created from $\cX^t$.
\emph{Exemplar} images in $\cP^t$ are the selected subset of images used in training future tasks. 
Thus, training at task $t$ uses images $\cX^t$ and $\cP^i$, where $i<t$.

Training on this additional old class data can help mitigate the effect of catastrophic forgetting for previously seen classes.
In iCARL~\cite{RKS+17} the exemplar selection used to create $\cP^t$ is done such that the selected set of exemplars should approximate the class mean vector well,
using a \emph{herding} algorithm~\cite{W09}. 
Such an approach can bound the memory requirement for stored examples. 
\section{Memory-efficient incremental learning}
\label{sec:method}

Our goal is to preserve compact feature descriptors, \ie $\vv \defn h\prm{\theta}(x)$, instead of images from old classes. This enables us to be significantly more memory-efficient, or to store more examples per class given the same memory requirement.

The major challenge of preserving only the feature descriptors is that it is not clear how they would evolve over time as the feature extractor $h\prm{\theta}$ is trained on new data. 
This introduces a problem for the new tasks, where we would like to use all preserved feature descriptors to learn a feature classifier $g\prm{\tilde{W}}$ on all classes jointly.
Preserved feature descriptors are \emph{not compatible} with feature descriptors from the new task because $h\prm{\theta}$ is different.
Furthermore, we cannot re-extract feature descriptors from $h\prm{\theta}$ if we do not have access to old images.

We propose a \emph{feature adaptation} process, which directly updates the feature descriptors as the network changes with a \emph{feature adaptation network} $\phi\prm{\psi}$.
During training of each task, we first train the CNN using classification and distillation losses (Sec.~\ref{sec:extra}).
Then, we learn the feature adaptation network (Sec.~\ref{sec:featureadapt}) and use it to adapt stored features from previous tasks to the current feature space.
Finally, a feature classifier $g\prm{\tilde{W}}$ is learned with features from both the current task and the adapted features from the previous tasks (Sec.~\ref{sec:finetuning}).
This feature classifier $g\prm{\tilde{W}}$ is used to classify the features extracted from test images and is independent from the network classifier $g\prm{W}$, which is used to train the network.
Figure~\ref{fig:overview} gives a visual overview of our approach (see also Algorithm 1 in Appendix A). We describe it in more detail in the following.

\subsection{Network training}
\label{sec:extra}

This section describes the training of the backbone convolutional neural network $f\prm{\theta,W}$.
Our implementation follows the same training setup as in Section~\ref{sec:background} with two additional components: \textit{cosine normalization} and \textit{feature distillation}.

\head{Cosine normalization}
was proposed in various learning tasks~\cite{LH17,LZX+18}, including incremental learning~\cite{HPL+19}.
The prediction of the network~\eqref{equ:netOutput} is based on cosine similarity, instead of simple dot product.
This is equivalent to $\hat{W}\tran \hat{\vv}$, where $\hat{W}$ is the column-wise $\ell_2$-normalized counterpart of parameters $W$ of the classifier, and $\hat{\vv}$ is the $\ell_2$-normalized counterpart of the feature $\vv$.

\head{Feature distillation}
is an additional distillation term based on feature descriptors instead of logits.
Similar to~\eqref{eq:ldiss}, we add a constraint in the loss function which encourages the new feature extractor $h\prm{\theta}^t$ to mimic the old one $h\prm{\theta}^{t-1}$:
\begin{equation}
L_{\text{FD}}(x) \defn 1 - \text{cos}(h\prm{\theta}^{t}(x),h\prm{\theta}^{t-1}(x)),
\label{eq:fdiss}
\end{equation}
where $x \in \cX^t$ and $h\prm{\theta}^{t-1}$ is the frozen feature extractor from the previous task.
The feature distillation loss term is minimized together with other loss terms,
\begin{equation}
L(x,y) \defn L_{\text{CE}}(x,y) + \lambda L_{\text{KD}}(x) + \gamma L_{\text{FD}}(x),
\label{eq:lcetwodist}
\end{equation}
where $\gamma$ is a tuned hyper-parameter. We study its impact in Section~\ref{sec:val}.

Feature distillation has already been applied in incremental learning as a replacement for the knowledge distillation loss~\eqref{eq:ldiss}, but only to the feature vectors of preserved images~\cite{HPL+19}.
It is also similar in spirit to \emph{attention distillation}~\cite{DSP+19}, which adds a constraint on the attention maps produced by the two models. 

Cosine normalization and feature distillation improve the accuracy of our method and the baselines. 
The practical impact of these components will be studied in more detail in Section~\ref{sec:exp}.

\subsection{Feature adaptation}
\label{sec:featureadapt}
\head{Overview.}
Feature adaptation is applied after CNN training at each task.
We first describe feature adaptation for the initial two tasks and then extend it to subsequent tasks.
At task $t=1$ , the network is trained with images $\cX^1$ belonging to classes $\cC^1$.
After the training is complete, we extract feature descriptors 
$\cV^1 =\{(h\prm{\theta}^1(x) | x \in \cX^1\}$, where $h\prm{\theta}^1(x)$ refers to the feature extractor component of $f\prm{\theta,W}^1$.
We store these features in memory $\cM^1 = \cV^1$ after the first task\footnote{We also store the corresponding label information.}. 
We also reduce the number of features stored in $\cM^1$ to fit specific memory requirements, which is explained later in the section.
At task $t=2$, we have a new set of images $\cX^2$ belonging to new classes $\cC^2$.
The network $f\prm{\theta,W}^2$ is initialized from $f\prm{\theta,W}^1$, where $f\prm{\theta,W}^1$ is fixed and kept as reference during training with distillation~\eqref{eq:lcetwodist}. 
After the training finishes, we extract features  
$\cV^2 = \{(h\prm{\theta}^2(x) | x \in \cX^2\}$.

We now have two sets of features, $\cM^1$ and $\cV^2$ extracted from two tasks that correspond to different sets of classes.
Importantly, $\cM^1$ and $\cV^2$ are extracted with different feature extractors, $h\prm{\theta}^1$ and $h\prm{\theta}^2$, respectively.
Hence, the two sets of vectors lie in different feature spaces and are not compatible with each other.
Therefore, we must transform features $\cM^1$ to the same feature space as $\cV^2$.
We train a feature adaptation network $\phi\prm{\psi}^{1 \rightarrow 2}$ to map $\cM^1$ to the same space as $\cV^2$ (training procedure described below).

Once the feature adaptation network is trained, we create a new memory set $\cM^2$ by transforming the existing features in the memory $\cM^1$ to the same feature space as $\cV^2$, \ie $\cM^2 = \cV^2 \cup \phi\prm{\psi}^{1 \rightarrow 2}(\cM^1)$.
The resulting $\cM^2$ contains new features from the current task and adapted features from the previous task, and can be used to learn a discriminative feature classifier explained in Section~\ref{sec:finetuning}.
$\cM^1$ and $f\prm{\theta,W}^1$ are no longer stored for future tasks.

We follow the same procedure for subsequent tasks $t>2$. We have a new set of data with images $\cX^t$ belonging to classes $\cC^t$.  
Once the network training is complete after task $t$, we extract features descriptors 
$\cV^t = \{(h\prm{\theta}^t(x) | x \in \cX^t\}$.
We train a feature adaptation network $\phi\prm{\psi}^{(t-1) \rightarrow t}$ 
and use it to create $\cM^t = \cV^t \cup \phi\prm{\psi}^{(t-1) \rightarrow t}(\cM^{t-1})$.
The memory set $\cM^t$ will have features stored from \emph{all classes} $\cC^i$, $i \leq t$, transformed to the current feature space of $h\prm{\theta}^t$. 
$\cM^{t-1}$ and $f\prm{\theta,W}^{t-1}$ are no longer needed for future tasks.

\head{Training the feature adaptation network $\phi\prm{\psi}$.}
At task $t$, we transform $\cV^{t-1}$ to the same feature space as $\cV^{t}$.
We do this by learning a transformation function $\phi\prm{\psi}^{(t-1) \rightarrow t}: \real^d \rightarrow \real^d$, that maps output of the previous feature extractor $h\prm{\theta}^{t-1}$ to the current feature extractor $h\prm{\theta}^{t}$ using the current task images $\cX^t$.

Let ${\cV}^{'} = \{(h\prm{\theta}^{t-1}(x),h\prm{\theta}^{t}(x))| x \in \cX^t\}$ and  $(\overline{\vv},\vv) \in \cV^{'}$.
In other words, given an image $x \in \cX^{t}$, $\overline{\vv}$ corresponds to its feature extracted with $h\prm{\theta}^{t-1}(x)$, the state of the feature extractor after task $t-1$.
On the other hand, $\vv$ corresponds to the feature representation of the same image $x$, but extracted with the model at the end of the current task, \ie $h\prm{\theta}^t(x)$.
Finding a mapping between $\overline{\vv}$ and $\vv$ allows us to map other features in $\cM^{t-1}$ to the same feature space as $\cV^t$.

When training the \emph{feature adaptation network} $\phi\prm{\psi}^{(t-1) \rightarrow t}$, we use a similar loss function as the feature hallucinator~\cite{HG17}:
\begin{equation}
L_{\text{fa}}(\overline{\vv},\vv,y) \defn \alpha L_{\text{sim}}(\vv,\phi\prm{\psi}(\overline{\vv})) + L_{\text{cls}}(g\prm{W},\phi\prm{\psi}(\overline{\vv}),y),
\label{eq:adapttrain}
\end{equation}
where $y$ is the corresponding label to $\vv$.
The first term $L_{\text{sim}}(\vv,\phi\prm{\psi}(\overline{\vv})) = 1 - \text{cos}(\vv,\phi\prm{\psi}(\overline{\vv}))$ encourages the adapted feature descriptor $\phi\prm{\psi}(\overline{\vv})$ to be similar to $\vv$, its counterpart extracted from the updated network.
Note that this is the same loss function as feature distillation~\eqref{eq:fdiss}. 
The purpose of this method is transforming features between different feature spaces, whereas feature distillation is helpful by preventing features from drifting too much in the feature space. The practical impact of feature distillation will be presented in more detail in Section~\ref{sec:val}.
The second loss term $L_{\text{cls}}(g\prm{W},\phi\prm{\psi}(\overline{\vv}),y)$ is the cross-entropy loss and $g\prm{W}$ is the \emph{fixed} network classifier of the network $f\prm{\theta,W}$.
This term encourages adapted feature descriptors to belong to the correct class $y$.

\head{Reducing the size of $\cM^t$.}
The number of stored vectors in memory $\cM^t$ can be reduced to satisfy specified memory requirements.
We reduce the number of features in the memory by \emph{herding}~\cite{RKS+17,W09}.
Herding is a greedy algorithm that chooses the subset of features that best approximates the class mean.
When updating the memory after task $t$, we use herding to only keep a fixed number ($L$) of features per class, \ie $\cM^t$ has $L$ vectors per class.

\subsection{Training the feature classifier $g\prm{\tilde{W}}$}
\label{sec:finetuning}
Our goal is to classify unseen test images belonging to $K^t = \sum_{i=1}^t |\cC^i|$ classes,  which includes classes from previously seen tasks.
As explained in Sec.~\ref{sec:background}, the learned network $f\prm{\theta,W}^t$ is a mapping from images to $K^t$ classes and can be used to classify test images.
However, training only on $\cX^t$ images results in  sub-optimal performance, because the previous tasks are still forgotten to an extent, even when using distillation~\eqref{eq:fdiss} during training.
We leverage the preserved \emph{adapted} feature descriptors from previous tasks to learn a more accurate feature classifier.

At the end of task $t$, a new feature classifier $g\prm{\tilde{W}}^{t}$ is trained with the memory $\cM^t$, which contains the adapted feature descriptors from previous tasks as well as feature descriptors from the current task.
This is different than the network classifier $g\prm{W}^{t}$, which is a part of the network $f\prm{\theta,W}^t$.
When given a test image, we extract its feature representation with $h\prm{\theta}^{t}$ and classify it using the feature classifier $g\prm{\tilde{W}}^{t}$.
In practice, $g\prm{\tilde{W}}^{t}$ is a linear classifier which can be trained in various ways, \eg linear SVM, SGD \etc We use Linear SVMs in our experiments.

\section{Experiments}
\label{sec:exp}

We describe our experimental setup, then show our results on each dataset in terms of classification accuracy. 
We also measure the quality of our feature adaptation method, which is independent of the classification task.
Finally, we study in detail the impact of key implementation choices and parameters.

\subsection{Experimental setup}
\label{sec:setup}

\head{Datasets.}
We use CIFAR-100~\cite{KH09}, ImageNet-100 and ImageNet-1000 in our experiments.
ImageNet-100~\cite{RKS+17} is a subset of the ImageNet-1000 dataset~\cite{RDH+15}  
containing $100$ randomly sampled classes from the original $1000$ classes. 
We follow the same setup as iCARL~\cite{RKS+17}.
The network is trained in a class-incremental way, only considering the data available at each task.
We denote the number of classes at each task by $M$, and total number of tasks by $T$. 
After each task, classification is performed on all classes seen so far. 
Every CIFAR-100 and ImageNet-100 experiment was performed 5 times with random class orderings. Reported results are averaged over all 5 runs.

Two evaluation metrics are reported.
The first is a curve of classification accuracies on all classes that have been trained after each task.
The second is the \emph{average incremental accuracy}, which is the average of points in first metric.
Top-$1$ and top-$5$ accuracy is computed for CIFAR-100 and ImageNet respectively. 

\head{Baselines.}
Our main baselines are given by the two methods in the literature that we extend. 
Learning Without Forgetting (LwF)~\cite{LH17} does not preserve any data from earlier tasks and is trained with classification and distillation loss terms~\eqref{eq:lcedist}. 
We use the multi-class version (LwF.MC) proposed by Rebuffi~\etal~\cite{RKS+17}.
iCARL \cite{RKS+17} extends LwF.MC by preserving representative training images of previously seen classes.
All experiments are reported with our implementation unless specified otherwise.
Rebuffi~\etal~\cite{RKS+17} fix the total number of exemplars stored at any point, and change the number of exemplars per class depending on the total number of classes.
Unlike the original iCARL, we fix the number of exemplars per class as $P$ in our implementation (as in~\cite{HPL+19}).
We extend the original implementations of iCARL and LwF by applying cosine normalization and feature distillation loss (see Sec.~\ref{sec:extra}), as these variants have shown to improve the accuracy.
We refer to the resulting variants as $\gamma$-iCARL and $\gamma$-LwF respectively ($\gamma$ is the parameter that controls the feature distillation 
in Eq.~\eqref{eq:fdiss}).

\head{Implementation details.}
The feature extraction network $h\prm{\theta}$ is Resnet-32~\cite{HZRS16} ($d=64$) for CIFAR100 and Resnet-18~\cite{HZRS16} ($d=512$) for ImageNet-100 and ImageNet-1000. We use a Linear SVM~\cite{CV95,scikit-learn} for our feature classifier $g\prm{\tilde{W}}$.
The feature adaptation network $\phi\prm{\psi}$ is a 2-layer multilayer perceptron (MLP) with ReLU~\cite{GBB11} activations and $d$ input/output and $d'=16d$ hidden dimensions.
We use binary cross-entropy for the loss function~\eqref{eq:lcedist}, and $\lambda$ for the knowledge distillation~\eqref{eq:lcedist} is set to $1$.
Consequently, the activation function $\vsigma$ is sigmoid.
We use the same hyper-parameters as Rebuffi~\etal~\cite{RKS+17} when training the network,
a batch size of $128$, weight decay of $1\mathrm{e}{-5}$, and learning rate of $2.0$.
In CIFAR-100, we train the network for $70$ epochs at each task, and reduce the learning rate by a factor of $5$ at epochs $50$ and $64$.
For ImageNet experiments, we train the network for $60$ epochs at each task, and reduce the learning rate by a factor of $5$ at epochs $20$, $30$, $40$ and $50$.

\subsection{Impact of memory footprint}
\label{sec:memory}
\begin{figure*}[t]
\input{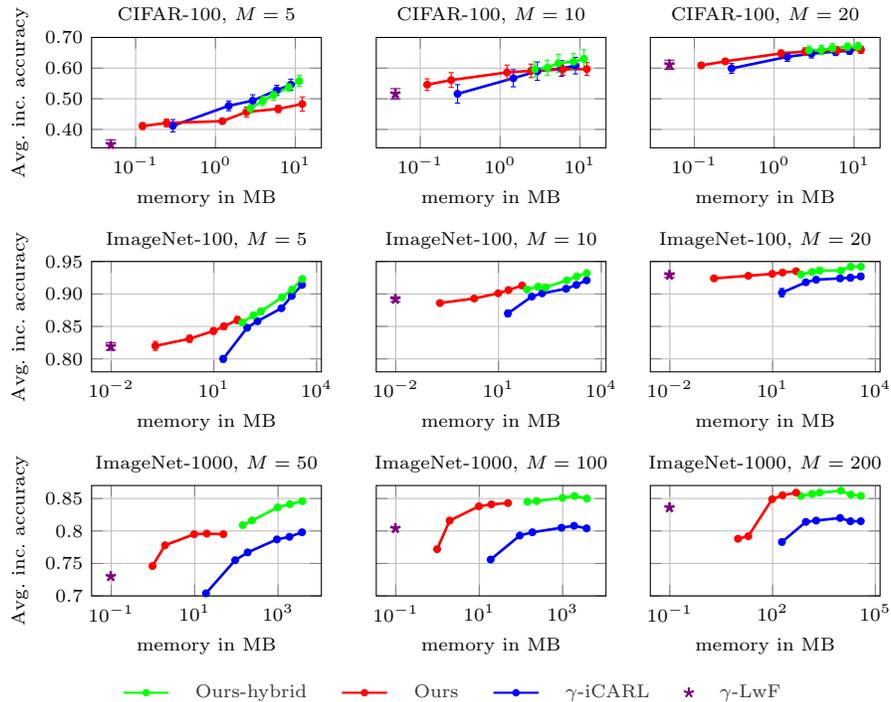}
\caption{Memory (in MB) vs average incremental accuracy on CIFAR-100, ImageNet-100 and ImageNet-1000 for different number of classes per task ($M$). We vary the memory requirement for our method and $\gamma$-iCARL by changing the number of preserved feature descriptors ($L$) and images ($P$) respectively. For Ours-hybrid, we set $L=100$ (CIFAR) and $L=250$ (ImageNet-100 and ImageNet-1000) and vary $P$.
\label{fig:mem_all}
}
\end{figure*}

Our main goal is to improve the memory requirements of an incremental learning framework. 
We start by comparing our method against our baselines in terms of memory footprint.
Figure~\ref{fig:mem_all} shows the memory required by each method and the corresponding average incremental accuracy.
The memory footprint is all the preserved data (features or images) for all classes of the dataset.
Memory footprint for $\gamma$-iCARL is varied by changing $P$, the fixed number of images preserved for each class.
Memory footprint for our method is varied by changing $L$, the fixed number of feature descriptors per class (Sec.~\ref{sec:featureadapt}).
We also present \emph{Ours-hybrid}, a variant of our method where we keep $P$ images and $L$ feature descriptors. In this variant, we vary $P$ to fit specified memory requirements.

Figure~\ref{fig:mem_all} shows average incremental accuracy for different memory usage on CIFAR-100, ImageNet-100 and ImageNet-1000.
Note that while our method still achieves higher or comparable accuracy compared on CIFAR-100, the memory savings are less significant. 
That is due to the fact that images have lower resolution ($32\times32\times3$ uint8, 3.072KB) and preserving feature descriptors ($d=64$ floats, 0.256KB) has less impact on the memory in that dataset. However, due to the lower computation complexity of training on CIFAR-100, we use CIFAR-100 to tune our hyperparameters (Sec.~\ref{sec:val}).
The memory savings with our method are more significant for ImageNet.
The resolution of each image in ImageNet is $256\times256\times3$, \ie,
storing a single uint8 image in the memory takes $192$ KB.
Keeping a feature descriptor of $d=512$ floats is significantly cheaper; it only requires $2$ KB.
This is about $\sim1\%$ of the memory required for an image.
Note there are many compression techniques for both images and features (\eg JPEG, HDF5, PCA). Our analysis will solely focus on uncompressed data.

We achieve the same accuracy with significantly less memory compared to $\gamma$-iCARL on ImageNet datasets.
The accuracy of our method is superior to $\gamma$-iCARL when $M \geq 100$ on ImageNet-1000.
Memory requirements are at least an order of magnitude less in most cases.
Ours-hybrid shows that we can preserve features with smaller number of images and further improve accuracy.
This results in higher accuracy compared to $\gamma$-iCARL for the same memory footprint.

\begin{figure*}[t]
\newenvironment{customlegend}[1][]{%
    \begingroup
    \csname pgfplots@init@cleared@structures\endcsname
    \pgfplotsset{#1}%
}{%
    \csname pgfplots@createlegend\endcsname
    \endgroup
}%
\def\addlegendimage{\csname pgfplots@addlegendimage\endcsname}

\input{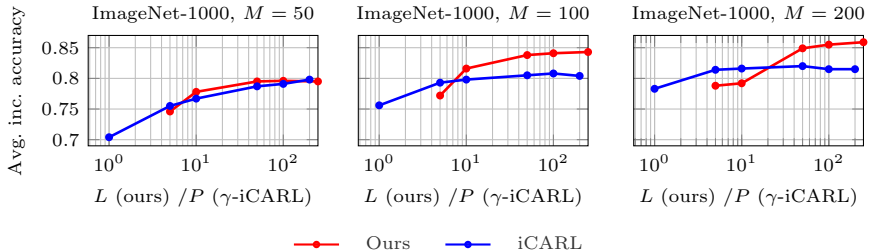}
\begin{tabular}{@{\xssp}c@{\xssp}c@{\xssp}c@{\xssp}}
{
\begin{tikzpicture}
\begin{axis}[%
  width=0.38\linewidth,
  height=0.25\linewidth,
	xlabel={$L$ (ours) /$P$ ($\gamma$-iCARL)},
   	xmax = 250,
   	xtick={0.1, 1, 10,100}, % only way to get rid of minor ticks...
   	xmode=log,
    grid=both,
	ylabel= Avg. inc. accuracy,
	xminorticks=false,
  title= {ImageNet-1000, $M=50$},
    ymin = 0.69,
    ymax = 0.87,
	legend cell align={left},
	legend pos=outer north east,
    legend style={at={(0.5,1.5)},anchor=north, font =\scriptsize, fill opacity=0.8, row sep=-2.5pt, legend columns=-1},
  	title style={yshift=-1.1ex}
]

	\addplot[color=red,     solid, mark=*,  mark size=1.0, line width=1.0, error bars/.cd, y dir=both, y explicit] table[x=L, y expr={\thisrow{ours}}, y error expr={\thisrow{ours-err}/1}] \memImnetBigFifty;
	\addplot[color=blue,     solid, mark=*,  mark size=1.0, line width=1.0, error bars/.cd, y dir=both, y explicit] table[x=L, y expr={\thisrow{icarl}}, y error expr={\thisrow{icarl-err}/1}] \memImnetBigFifty;

\end{axis}
\end{tikzpicture}
}
&
{
\begin{tikzpicture}
\begin{axis}[%
  width=0.38\linewidth,
  height=0.25\linewidth,
  xlabel={$L$ (ours) /$P$ ($\gamma$-iCARL)},
    xmax = 250,
    xtick={0.1, 1, 10,100}, % only way to get rid of minor ticks...
   	xmode=log,
    grid=both,
    ymin = 0.69,
    ymax = 0.87,
  yticklabels={,,},
	xminorticks=false,
  title= {ImageNet-1000, $M=100$},
  	title style={yshift=-1.1ex}
]

	\addplot[color=red,     solid, mark=*,  mark size=1.0, line width=1.0, error bars/.cd, y dir=both, y explicit] table[x=L, y expr={\thisrow{ours}}, y error expr={\thisrow{ours-err}/1}] \memImnetBigOneHundred;
	\addplot[color=blue,     solid, mark=*,  mark size=1.0, line width=1.0, error bars/.cd, y dir=both, y explicit] table[x=L, y expr={\thisrow{icarl}}, y error expr={\thisrow{icarl-err}/1}] \memImnetBigOneHundred;

\end{axis}
\end{tikzpicture}
}
&
{
\begin{tikzpicture}
\begin{axis}[%
  width=0.38\linewidth,
  height=0.25\linewidth,
  xlabel={$L$ (ours) /$P$ ($\gamma$-iCARL)},
    xmax = 250,
    xtick={0.1, 1, 10,100}, % only way to get rid of minor ticks...
   	xmode=log,
    grid=both,
    ymin = 0.69,
    ymax = 0.87,
    yticklabels={,,},
	title= {ImageNet-1000, $M=200$},
	title style={yshift=-1.1ex}
]

	\addplot[color=red,     solid, mark=*,  mark size=1.0, line width=1.0, error bars/.cd, y dir=both, y explicit] table[x=L, y expr={\thisrow{ours}}, y error expr={\thisrow{ours-err}/1}] \memImnetBigTwoHundred;
	\addplot[color=blue,     solid, mark=*,  mark size=1.0, line width=1.0, error bars/.cd, y dir=both, y explicit] table[x=L, y expr={\thisrow{icarl}}, y error expr={\thisrow{icarl-err}/1}] \memImnetBigTwoHundred;

\end{axis}
\end{tikzpicture}
}
\end{tabular}

\centering
\begin{tikzpicture}
\begin{customlegend}[legend columns=5,legend style={align=center,draw=none,column sep=2ex},
        legend entries={
                        {\scriptsize Ours} ,
                        {\scriptsize iCARL} ,
                        }]
        \addlegendimage{color=red, solid, mark=*, mark size=1.0}   
        \addlegendimage{color=blue, solid, mark=*, mark size=1.0}   
        \end{customlegend}
\end{tikzpicture}
\vspace{-10pt}
\caption{Impact of $L$, the number of features stored per class for ours, and $P$, the number of images stored per class for $\gamma$-iCARL. $M$ is the number of classes per task.
\label{fig:L_all}
}
\end{figure*}

Figure~\ref{fig:L_all} shows the accuracy for different number of preserved \emph{data points} on ImageNet-1000, where \emph{data points} refer to features for our method and images for $\gamma$-iCARL.
Our method outperforms $\gamma$-iCARL in most cases, even if ignoring the memory savings of features compared to images.

\subsection{Comparison to state of the art}
\label{sec:sota}

\begin{figure*}[b]
\newenvironment{customlegend}[1][]{%
    \begingroup
    \csname pgfplots@init@cleared@structures\endcsname
    \pgfplotsset{#1}%
}{%
    \csname pgfplots@createlegend\endcsname
    \endgroup
}%
\def\addlegendimage{\csname pgfplots@addlegendimage\endcsname}

\input{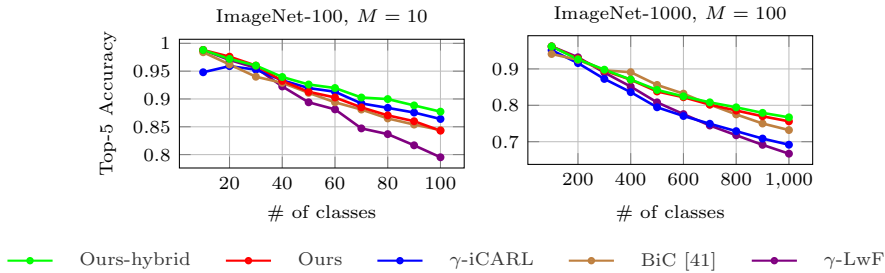}
\centering
\begin{tabular}{@{\xssp}c@{\xssp}c@{\xssp}}
{
\begin{tikzpicture}
\begin{axis}[%
	width=0.44\linewidth,
  height=0.27\linewidth,
	xlabel={\# of classes},
   	% xmin = 0.1,
   	% xmax = 200,
   	% xtick={0.1, 1, 10,100,201}, % only way to get rid of minor ticks...
   	% xticklabels={$10^0$,$10^1$,$10^2$,$10^3$,$10^4$,$10^5$},
   	% xmode=log,
    grid=both,
	ylabel= Top-5 Accuracy,
	xminorticks=false,
	title= {ImageNet-100, $M=10$},
	legend cell align={left},
	legend pos=outer north east,
    legend style={at={(0.5,1.5)},anchor=north, font =\scriptsize, fill opacity=0.8, row sep=-2.5pt, legend columns=-1},
  	title style={yshift=-1.1ex}
]

	% \addplot[color=brown, solid, mark=*, mark size=1.0, line width = 1.0, error bars/.cd, y dir=both, y explicit] table[x=kb, y expr={\thisrow{noadapt}}, y error expr={\thisrow{noadapt-err}}] \memCifarTen;
	\addplot[color=violet,     solid, mark=*,  mark size=1.0, line width=1.0, error bars/.cd, y dir=both, y explicit] table[x=classes, y expr={\thisrow{lwf}}] \taskImnetTen;
	\addplot[color=blue,     solid, mark=*,  mark size=1.0, line width=1.0, error bars/.cd, y dir=both, y explicit] table[x=classes, y expr={\thisrow{icarl}}] \taskImnetTen;
	\addplot[color=brown,     solid, mark=*,  mark size=1.0, line width=1.0, error bars/.cd, y dir=both, y explicit] table[x=classes, y expr={\thisrow{bias}}] \largescalehundred;
	\addplot[color=red,     solid, mark=*,  mark size=1.0, line width=1.0, error bars/.cd, y dir=both, y explicit] table[x=classes, y expr={\thisrow{ours}}] \taskImnetTen;
	\addplot[color=green,     solid, mark=*,  mark size=1.0, line width=1.0, error bars/.cd, y dir=both, y explicit] table[x=classes, y expr={\thisrow{oursicarl}}] \taskImnetTen;

\end{axis}
\end{tikzpicture}
}
&
{
\begin{tikzpicture}
\begin{axis}[%
  width=0.44\linewidth,
  height=0.27\linewidth,
	xlabel={\# of classes},
   	% xmin = 0.1,
   	% xmax = 200,
   	% xtick={0.1, 1, 10,100,201}, % only way to get rid of minor ticks...
   	% xticklabels={$10^0$,$10^1$,$10^2$,$10^3$,$10^4$,$10^5$},
   	% xmode=log,
    grid=both,
% 	ylabel= Top-5 Accuracy,
	xminorticks=false,
  title= {ImageNet-1000, $M=100$},
	legend cell align={left},
	legend pos=outer north east,
    legend style={at={(0.5,1.5)},anchor=north, font =\scriptsize, fill opacity=0.8, row sep=-2.5pt, legend columns=-1},
  	title style={yshift=-1.1ex}
]

	% \addplot[color=brown, solid, mark=*, mark size=1.0, line width = 1.0, error bars/.cd, y dir=both, y explicit] table[x=kb, y expr={\thisrow{noadapt}}, y error expr={\thisrow{noadapt-err}}] \memCifarTen;
	\addplot[color=violet,     solid, mark=*,  mark size=1.0, line width=1.0, error bars/.cd, y dir=both, y explicit] table[x=classes, y expr={\thisrow{lwf}}] \largescalethousand;
	\addplot[color=blue,     solid, mark=*,  mark size=1.0, line width=1.0, error bars/.cd, y dir=both, y explicit] table[x=classes, y expr={\thisrow{icarl}}] \largescalethousand;
	\addplot[color=brown,     solid, mark=*,  mark size=1.0, line width=1.0, error bars/.cd, y dir=both, y explicit] table[x=classes, y expr={\thisrow{bias}}] \largescalethousand;
	\addplot[color=red,     solid, mark=*,  mark size=1.0, line width=1.0, error bars/.cd, y dir=both, y explicit] table[x=classes, y expr={\thisrow{ours}}] \largescalethousand;
	\addplot[color=green,     solid, mark=*,  mark size=1.0, line width=1.0, error bars/.cd, y dir=both, y explicit] table[x=classes, y expr={\thisrow{oursicarl}}] \largescalethousand;

\end{axis}
\end{tikzpicture}
}

\end{tabular}

\centering
\begin{tikzpicture}
\begin{customlegend}[legend columns=5,legend style={align=center,draw=none,column sep=2ex},
        legend entries={{\scriptsize Ours-hybrid} ,
                        {\scriptsize Ours} ,
                        {\scriptsize $\gamma$-iCARL} ,
                        {\scriptsize BiC~\cite{WCW+19}} ,
                        {\scriptsize $\gamma$-LwF} ,
                        }]
        \addlegendimage{color=green, solid, mark=*, mark size=1.0}
        \addlegendimage{color=red, solid, mark=*, mark size=1.0}   
        \addlegendimage{color=blue, solid, mark=*, mark size=1.0}
        \addlegendimage{color=brown, solid, mark=*, mark size=1.0}
        \addlegendimage{color=violet, solid, mark=*, mark size=1.0}   
        \end{customlegend}
\end{tikzpicture}
\vspace{-10pt}
\caption{Classification curves of our method and state-of-the-art methods on ImageNet-100 and ImageNet-1000. $M$ is the number of classes per task.
\label{fig:sota}
}
\end{figure*}

Table~\ref{tab:sota_table} shows the total memory cost of the preserved data and average incremental accuracy of our method and existing works in the literature. 
Accuracy per task is shown in Figure~\ref{fig:sota}.
We report the average incremental accuracy for Ours when preserving $L=250$ features per class.
Ours-hybrid preserves $L=250$ features and $P=10$ images per class.
Baselines and state-of-the-art methods preserve $P=20$ images per class. 
It is clear that our method, which is the first work storing and adapting feature descriptors,
consumes significantly less memory than the other methods while improving the classification accuracy.

\begin{table*}

\centering
\scriptsize{
\begin{tabular}{@{\xssp}l@{\ssp}c@{\ssp}c@{\ssp}c@{\ssp}c@{\xssp}}
\toprule
             							& \multicolumn{2}{c}{ImageNet-100} 	 & \multicolumn{2}{c}{ImageNet-1000} \\
\midrule             		
										& Mem. in MB	& Accuracy		 & Mem. in MB	& Accuracy	  \\
\midrule             		
										& \multicolumn{4}{c@{\hspace{8mm}}}{\textsc{State-of-the-art methods}} \\

\midrule
Orig. LwF\cite{LH17}$^{\dagger*}$ 		& -	        	& 0.642          & -	 		& 0.566    \\ 
Orig. iCARL\cite{RKS+17}$^{\dagger*}$   & 375			& 0.836     	 & 3750			& 0.637	   \\
EEIL\cite{CMM+18}$^{\dagger}$			& 375			& 0.904			 & 3750			& 0.694    \\
Rebalancing\cite{HPL+19}                & 375           & 0.681          & 3750         & 0.643    \\
BiC w/o corr.\cite{WCW+19}$^{\dagger}$ 	& 375			& 0.872          & 3750			& 0.776    \\ 
BiC\cite{WCW+19}$^{\dagger}$        	& 375			& 0.906          & 3750			& 0.840    \\ 
\midrule
										& \multicolumn{4}{c@{\hspace{8mm}}}{\textsc{Baselines}} \\
\midrule
$\gamma$-LwF           					& -			    & 0.891          & -			& 0.804    \\ 
$\gamma$-iCARL        					& 375			& 0.914          & 3750			& 0.802    \\
\midrule
										& \multicolumn{4}{c@{\hspace{8mm}}}{\textsc{Our Method}} \\
\midrule
Ours         						    & \textbf{48.8}	& 0.913          & \textbf{488.3}& 0.843    			\\ 
Ours-hybrid 						    & 236.3		    & \textbf{0.927} & 2863.3		& \textbf{0.846}    	\\
\bottomrule
\\
\end{tabular}
}

\caption{Average incremental accuracy on ImageNet-100 with $M=10$ classes per task and ImageNet-1000 with $M=100$ classes per task. Memory usage shows the cost of storing images or feature vectors for all classes. $\dagger$ indicates that the results were reported from the paper. $*$ indicates numbers were estimated from figures in the paper.%\cite{RKS+17}.
\label{tab:sota_table}
}
\end{table*}

\begin{figure}[t]
\input{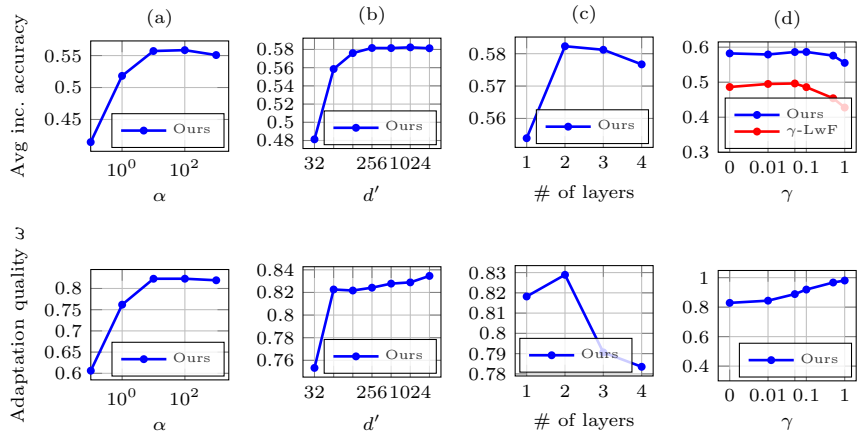}
\centering
% \resizebox{0.8\columnwidth}{!}{%
\begin{tabular}{@{\xssp}c@{\xssp}c@{\xssp}c@{\xssp}c@{\xssp}}
{
\begin{tikzpicture}
\begin{axis}[%
	width=0.28\linewidth,
	height=0.25\linewidth,
	xlabel={$\alpha$},
   	xmin = 0.1,
   	xmode=log,
    grid=both,
	legend cell align={left},
	legend pos=south east,
    legend style={cells={anchor=west}, font =\tiny, fill opacity=0.8, row sep=-2.5pt},
	ylabel= Avg inc. accuracy,
	xminorticks=false,
	title= {(a)},
  	title style={yshift=-1.1ex}
]
	\addplot[color=blue,     solid, mark=*,  mark size=1.0, line width=1.0] table[x=alpha, y expr={\thisrow{acc}}] \valAlpha;\leg{Ours};
\end{axis}
\end{tikzpicture}
}
&
{
\begin{tikzpicture}
\begin{axis}[%
	width=0.28\linewidth,
	height=0.25\linewidth,
	xlabel={$d'$},
   	xtick={32,64,128,256,512,1024,2048},
   	xticklabels={32,,,256,,1024,},
   	xmode=log,
	log basis x={2},
    grid=both,
	legend cell align={left},
	legend pos=south east,
    legend style={cells={anchor=west}, font =\tiny, fill opacity=0.8, row sep=-2.5pt},
	xminorticks=false,
	title= {(b)},
  	title style={yshift=-1.1ex}
]
	\addplot[color=blue,     solid, mark=*,  mark size=1.0, line width=1.0] table[x=hidden, y expr={\thisrow{acc}}] \valHidden;\leg{Ours};
\end{axis}
\end{tikzpicture}
}
&
{
\begin{tikzpicture}
\begin{axis}[%
	width=0.28\linewidth,
	height=0.25\linewidth,
	xlabel={\# of layers},
   	% xmin = 0.1,
   	% xtick={1,10,100,1000,10000,100000},
   	% xticklabels={$10^0$,$10^1$,$10^2$,$10^3$,$10^4$,$10^5$},
   	% xmode=log,
    grid=both,
	legend cell align={left},
	legend pos=south east,
    legend style={cells={anchor=west}, font =\tiny, fill opacity=0.8, row sep=-2.5pt},
	% ylabel= Avg incremental accuracy,
	xminorticks=false,
	title= {(c)},
  	title style={yshift=-1.1ex}
]
	\addplot[color=blue,     solid, mark=*,  mark size=1.0, line width=1.0] table[x=layers, y expr={\thisrow{acc}}] \valLayers;\leg{Ours};
\end{axis}
\end{tikzpicture}
}
&
{
\begin{tikzpicture}
\begin{axis}[%
	width=0.28\linewidth,
	height=0.25\linewidth,
	xlabel={$\gamma$},
   	xtick={0.001,0.01,0.05,0.1,1},
   	xticklabels={$0$,$0.01$,,$0.1$,$1$},
   	ymin=0.3,
   	xmode=log,
    grid=both,
	legend cell align={left},
	legend pos=south east,
    legend style={cells={anchor=west}, font =\tiny, fill opacity=0.8, row sep=-2.5pt},
	title= {(d)},
	title style={yshift=-1.1ex}
]

	\addplot[color=blue,     solid, mark=*,  mark size=1.0, line width=1.0] table[x=lambda, y expr={\thisrow{lwffeatgen}}] \valFD; \leg{Ours};
	\addplot[color=red,     solid, mark=*,  mark size=1.0, line width=1.0] table[x=lambda, y expr={\thisrow{lwf}}] \valFD; \leg{$\gamma$-LwF};

\end{axis}
\end{tikzpicture}
}
\\
{
\begin{tikzpicture}
\begin{axis}[%
	width=0.28\linewidth,
	height=0.25\linewidth,
	xlabel={$\alpha$},
   	xmin = 0.1,
   	xmode=log,
    grid=both,
	legend cell align={left},
	legend pos=south east,
    legend style={cells={anchor=west}, font =\tiny, fill opacity=0.8, row sep=-2.5pt},
	ylabel={Adaptation quality $\omega$},
	xminorticks=false,
  	title style={yshift=-1.1ex}
]
	\addplot[color=blue,     solid, mark=*,  mark size=1.0, line width=1.0] table[x=alpha, y expr={\thisrow{acc}}] \valAlphaQual;\leg{Ours};
\end{axis}
\end{tikzpicture}
}
&
{
\begin{tikzpicture}
\begin{axis}[%
	width=0.28\linewidth,
	height=0.25\linewidth,
	xlabel={$d'$},
   	xtick={32,64,128,256,512,1024,2048},
   	xticklabels={32,,,256,,1024,},
   	xmode=log,
	log basis x={2},
    grid=both,
	legend cell align={left},
	legend pos=south east,
    legend style={cells={anchor=west}, font =\tiny, fill opacity=0.8, row sep=-2.5pt},
	xminorticks=false,
  	title style={yshift=-1.1ex}
]
	\addplot[color=blue,     solid, mark=*,  mark size=1.0, line width=1.0] table[x=hidden, y expr={\thisrow{acc}}] \valHiddenQual;\leg{Ours};
\end{axis}
\end{tikzpicture}
}
&
{
\begin{tikzpicture}
\begin{axis}[%
	width=0.28\linewidth,
	height=0.25\linewidth,
	xlabel={\# of layers},
   	% xmin = 0.1,
   	% xtick={1,10,100,1000,10000,100000},
   	% xticklabels={$10^0$,$10^1$,$10^2$,$10^3$,$10^4$,$10^5$},
   	% xmode=log,
    grid=both,
	legend cell align={left},
	legend pos=south west,
    legend style={cells={anchor=west}, font =\tiny, fill opacity=0.8, row sep=-2.5pt},
	xminorticks=false,
  	title style={yshift=-1.1ex}
]
	\addplot[color=blue,     solid, mark=*,  mark size=1.0, line width=1.0] table[x=layers, y expr={\thisrow{acc}}] \valLayersQual;\leg{Ours};
\end{axis}
\end{tikzpicture}
}
&
{
\begin{tikzpicture}
\begin{axis}[%
	width=0.28\linewidth,
	height=0.25\linewidth,
	xlabel={$\gamma$},
   	xtick={0.001,0.01,0.05,0.1,1},
   	xticklabels={$0$,$0.01$,,$0.1$,$1$},
   	ymin=0.3,
   	xmode=log,
    grid=both,
	legend cell align={left},
	legend pos=south east,
    legend style={cells={anchor=west}, font =\tiny, fill opacity=0.8, row sep=-2.5pt},
	title style={yshift=-1.1ex}
]

	\addplot[color=blue,     solid, mark=*,  mark size=1.0, line width=1.0] table[x=lambda, y expr={\thisrow{ours}}] \valFDQual; \leg{Ours};
\end{axis}
\end{tikzpicture}
}

\end{tabular}%
% }
\caption{Impact of different parameters in terms of classification accuracy (top) and adaptation quality $\omega$ as defined in Sec.~\ref{sec:val} (bottom) on CIFAR-100: (a) similarity coefficient $\alpha$~\eqref{eq:adapttrain}, (b) size of hidden layers $d'$ of the feature adaptation network, (c) number of hidden layers in the feature adaptation network, (d) feature distillation coefficient $\gamma$~\eqref{eq:fdiss}.
\label{fig:val_cifar}
}
\end{figure}

\subsection{Impact of Parameters}
\label{sec:val}
We show the impact of the hyper-parameters of our method.
All experiments in this section are performed on a validation set created by holding out $10\%$ of the original CIFAR-100 training data.

\head{Impact of cosine classifier}
is evaluated on the base network, \ie LwF.MC.
We achieve $48.7$ and $45.2$ accuracy with and without cosine classifier respectively. 
We include cosine classifier in all baselines and methods.

\head{Impact of $\alpha$.}
The parameter $\alpha$ controls the importance of the similarity term \wrt classification term when learning the feature adaptation network~\eqref{eq:adapttrain}.
Figure~\ref{fig:val_cifar} top-(a) shows the accuracy with different $\alpha$.
The reconstruction constraint controlled by $\alpha$ requires a large value. 
We set $\alpha = 10^2$ in our experiments.

\head{Impact of $d'$.}
We evaluate the impact of $d'$, the dimensionality of the hidden layers of feature adaptation network $\phi\prm{\psi}$ in Figure~\ref{fig:val_cifar} top-(b).
Projecting feature vectors to a higher dimensional space is beneficial,
achieving the maximum validation accuracy with $d'= 1,024$.
We set $d'=16d$ in our experiments.

\head{Impact of the network depth.}
We evaluate different number of hidden layers of the feature adaptation network $\phi\prm{\psi}$ in Figure~\ref{fig:val_cifar} top-(c). 
The accuracy reaches its peak with two hidden layers, and starts to decrease afterwards, probably because the networks starts to overfit.
We use two hidden layers in our experiments.

\head{Impact of feature distillation.}
We evaluate different $\gamma$ for feature distillation~\eqref{eq:fdiss} for $\gamma$-LwF and our method, see Figure~\ref{fig:val_cifar} top-(d).
We set $\gamma=0.05$ and include feature distillation in all baselines and methods in our experiments.

\head{Quality of feature adaptation.}
We evaluate the quality of our feature adaptation process by measuring the average similarity between the adapted features and their \emph{ground-truth value}.
The ground-truth vector $h\prm{\theta}^t(x)$ for image $x$ is its feature representation if we actually had access to that image in task $t$.
We compare it against $\vv$, the corresponding vector of $x$ in the memory, that has been adapted over time.
We compute the feature adaptation quality by dot product $\omega = \vv\tran h\prm{\theta}^t(x)$.
This measures how accurate our feature adaptation is compared to the real vector if we had access to image $x$~\footnote{ $x$ is normally not available in future tasks, we use it here for the ablation study.}.

We repeat the validation experiments, this time measuring average $\omega$ of all vectors instead of accuracy (Figure~\ref{fig:val_cifar} bottom row).
Top and bottom rows of Figure~\ref{fig:val_cifar} shows that most trends are correlated meaning better feature adaptation results in better accuracy.
One main exception is the behavior of $\gamma$ in feature distillation~\eqref{eq:fdiss}.
Higher $\gamma$ results in higher $\omega$ but lower classification accuracy.
This is expected, as high $\gamma$ forces the network to make minimal changes to its feature extractor between different tasks, making feature adaptation more successful, but feature representations less discriminative.

\head{Effect of balanced feature classifier.} Class-imbalanced training is shown to lead to biased predictions \cite{WCW+19}. We investigate this in Supplementary Section C. Our experiments show that balancing the number of instances per class leads to improvements in the accuracy when training the feature classifier $g\prm{\tilde{W}}$. 
\section{Conclusions}
We have presented a novel method for preserving feature descriptors instead of images in incremental learning.
Our method introduces a \emph{feature adaptation} function, which accurately updates the preserved feature descriptors as the network is updated with new classes.
The proposed method is thoroughly evaluated in terms of classification accuracy and adaptation quality, showing
that it is possible to achieve state-of-the-art  accuracy with a significantly lower memory footprint.
Our method is orthogonal to existing work~\cite{LH17,RKS+17} and can be combined to achieve even higher accuracy with low memory requirements.

\footnotesize{
\head{Acknowledgements.} This research was funded in part by NSF grants IIS 1563727 and IIS 1718221, Google Research Award, Amazon Research Award, and AWS Machine Learning Research Award.
}

\clearpage
\bibliographystyle{splncs04}
\bibliography{egbib}

\newpage
\appendix
\section{Algorithm}

An overview of our framework is described in Algorithm~\ref{alg:main}.

\begin{algorithm}
\caption{Memory-efficient incremental learning}
\label{alg:main}
\begin{algorithmic}[1]
\begin{scriptsize}
\Procedure{Algorithm}{Training examples $\cX$, labels $\cY$}
\State Given $T$ tasks
\State *** Train first task ***
\State $\cX^1, \cY^1 \in \cX, \cY$ 
 \algocom{Data examples for first task}
\State $\theta, W =$ \Call{Optimize}{$L_{CE}(f\prm{\theta,W}(\cX^1), \cY^1)$} 
 \algocom{(\ref{eq:lcedist})}
\State $h\prm{\theta}^1 = h\prm{\theta}$
 \algocom{Freeze feature extractor}
\State $\vM^1 = h\prm{\theta}^1(\cX^1)$
 \algocom{Store feature descriptors of images}
\State $\vM^1 =$ \Call{Herding}{$\vM^1$}
 \algocom{Reduce number of stored features}
\For{$t \in [2,\ldots, T]$}
\State *** Train incremental tasks ***
	\State $\cX^t, \cY^t \in \cX, \cY$ 
 	\algocom{Data examples for current task}
	\State $\theta, W =$ \Call{Optimize}{$L(f\prm{\theta,W}(\cX^t), \cY^t)$} 
	\algocom{(\ref{eq:lcetwodist})}
	\State $h\prm{\theta}^t = h\prm{\theta}$
    \State $\phi\prm{\psi} =$ \Call{Feature Adaptation}{$h\prm{\theta}^t, h\prm{\theta}^{t-1}, \cX^t, \cY^t$}
    \State $\vM^t = h^t\prm{\theta}(\cX^t)$ 
     \algocom{Store new feature descriptors}
    \State $\vM^t =$ \Call{Herding}{$\vM^t$}
    \State $\vM^t = \vM^{t} \cup \phi\prm{\psi}(\vM^{t-1})$
     \algocom{Adapt stored features}
    \State $\tilde{W} =$ \Call{Train Classifier}{$\vM^t, \cY^{1,\dots,t}$}
     \algocom{(Sec.~\ref{sec:finetuning})}

\EndFor
\EndProcedure
\end{scriptsize}
\end{algorithmic}
\begin{algorithmic}[1]
\begin{scriptsize}
\Procedure{FeatureAdaptation}{$h\prm{\theta}^{old}, h\prm{\theta}^{new}, \cX, \cY$}
\State *** Returns transformation function ***
\State $\overline{\vV} = h\prm{\theta}^{old}(\cX)$
	\algocom{Feature descriptors of old extractor}
\State $\vV = h\prm{\theta}^{new}(\cX)$
	\algocom{Feature descriptors of new extractor}
\State $\psi =$ \Call{Optimize}{$L_{\text{FA}}$($\overline{\vV}$, $\vV, \cY$)}
\algocom{(\ref{eq:adapttrain})}
\State \Return $\phi\prm{\psi}$
\EndProcedure
\end{scriptsize}
\end{algorithmic}

\end{algorithm}

\section{Feature Adaptation Quality}

\begin{figure}
\input{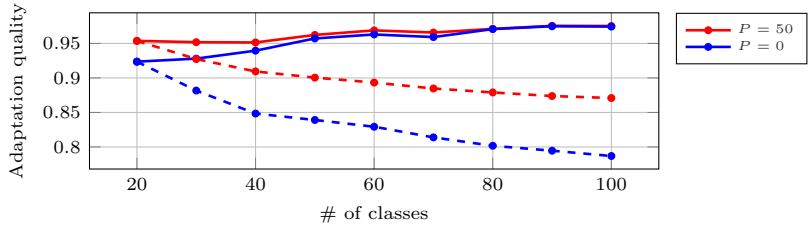}
\centering
% \resizebox{0.5\columnwidth}{!}{%
\begin{tabular}{@{\xxssp}c@{\xxssp}}
% {
\begin{tikzpicture}
\begin{axis}[%
	width=0.75\textwidth,
	height=0.3\textwidth,
	xlabel={\# of classes},
   	% xmin = 0.1,
   	% xtick={1,10,100,1000,10000,100000},
   	% xticklabels={$10^0$,$10^1$,$10^2$,$10^3$,$10^4$,$10^5$},
   	% xmode=log,
    grid=both,
	legend cell align={left},
	legend pos=outer north east,
    legend style={cells={anchor=west}, font =\tiny, fill opacity=0.8, row sep=-2.5pt},
	ylabel= {\scriptsize Adaptation quality},
	xminorticks=false,
  	title style={yshift=-1.1ex}
]

	% \addplot[color=blue,     solid, mark=*,  mark size=1.0, line width=1.0] table[x=task, y expr={\thisrow{twohundred}}] \genQualityLast;
	% \addplot[color=brown,     solid, mark=*,  mark size=1.0, line width=1.0] table[x=task, y expr={\thisrow{hundred}}] \genQualityLast;\leg{$\tiny P=100$};
	\addplot[color=red,     solid, mark=*,  mark size=1.0, line width=1.0] table[x=task, y expr={\thisrow{fifty}}] \genQualityLast;
	% \addplot[color=orange,     solid, mark=*,  mark size=1.0, line width=1.0] table[x=task, y expr={\thisrow{twenty}}] \genQualityLast;
	% \addplot[color=green,     solid, mark=*,  mark size=1.0, line width=1.0] table[x=task, y expr={\thisrow{ten}}] \genQualityLast;
	\addplot[color=blue,     solid, mark=*,  mark size=1.0, line width=1.0] table[x=task, y expr={\thisrow{zero}}] \genQualityLast;

	% \addplot[color=blue,     dashed, mark=*,  mark size=1.0, line width=1.0] table[x=task, y expr={\thisrow{twohundred}}] \genQualityFirst;\leg{$\small P=200$};
	% \addplot[color=brown,     solid, mark=*,  mark size=1.0, line width=1.0] table[x=task, y expr={\thisrow{hundred}}] \genQualityFirst;
	\addplot[color=red,     dashed, mark=*,  mark size=1.0, line width=1.0] table[x=task, y expr={\thisrow{fifty}}] \genQualityFirst;\leg{$\small P=50$};
	% \addplot[color=orange,     solid, mark=*,  mark size=1.0, line width=1.0] table[x=task, y expr={\thisrow{twenty}}] \genQualityFirst;\leg{$\scriptstyle K=20$};
	% \addplot[color=green,     dashed, mark=*,  mark size=1.0, line width=1.0] table[x=task, y expr={\thisrow{ten}}] \genQualityFirst;\leg{$\tiny P=10$};
	\addplot[color=blue,     dashed, mark=*,  mark size=1.0, line width=1.0] table[x=task, y expr={\thisrow{zero}}] \genQualityFirst;\leg{$\small P=0$};

\end{axis}
\end{tikzpicture}
% }
% &
% {
% \begin{tikzpicture}
% \begin{axis}[%
% 	width=0.21\textwidth,
% 	height=0.20\textwidth,
% 	xlabel={\# of classes},
%    	% xmin = 0.1,
%    	% xtick={1,10,100,1000,10000,100000},
%    	% xticklabels={$10^0$,$10^1$,$10^2$,$10^3$,$10^4$,$10^5$},
%    	% xmode=log,
%     grid=both,
	% legend cell align={left},
	% legend pos=outer north east,
 %    legend style={cells={anchor=west}, font =\tiny, fill opacity=0.8, row sep=-2.5pt},
% 	xminorticks=false,
% 	title= {First task only ($\omega_{1}$)},
%   	title style={yshift=-1.1ex},
% ]

% 	% \addplot[color=blue,     solid, mark=*,  mark size=1.0, line width=1.0] table[x=task, y expr={\thisrow{twohundred}}] \genQualityFirst;\leg{$\tiny P=200$};
% 	% % \addplot[color=brown,     solid, mark=*,  mark size=1.0, line width=1.0] table[x=task, y expr={\thisrow{hundred}}] \genQualityFirst;
% 	% \addplot[color=red,     solid, mark=*,  mark size=1.0, line width=1.0] table[x=task, y expr={\thisrow{fifty}}] \genQualityFirst;\leg{$\tiny P=50$};
% 	% % \addplot[color=orange,     solid, mark=*,  mark size=1.0, line width=1.0] table[x=task, y expr={\thisrow{twenty}}] \genQualityFirst;\leg{$\scriptstyle K=20$};
% 	% \addplot[color=green,     solid, mark=*,  mark size=1.0, line width=1.0] table[x=task, y expr={\thisrow{ten}}] \genQualityFirst;\leg{$\tiny P=10$};
% 	% \addplot[color=orange,     solid, mark=*,  mark size=1.0, line width=1.0] table[x=task, y expr={\thisrow{zero}}] \genQualityFirst;\leg{$\tiny P=0$};

% \end{axis}
% \end{tikzpicture}
% }
\end{tabular}%
% }
\caption{Feature adaptation quality on CIFAR-100 for $M=10$. $P$ refers to the number of images preserved in the memory.
Solid and dashed lines correspond to vectors from previous ($\omega_{t-1}$) and first ($\omega_{1}$) task respectively.
\label{fig:task_cifar_quality}
}
\vspace{-10pt}
\end{figure}

We evaluate the quality of our feature adaptation process by measuring the average similarity between the adapted features and their \emph{ground-truth value}.
We compute the feature adaptation quality as explained in Section~\ref{sec:val}.
However, we compute  two distinct measurements this time.
$\omega_{t-1}$ measures the average feature adaptation quality of features extracted in the previous task (\ie $y \in \cC^{t-1}$).
This measurement does not track the quality over time, but shows feature adaptation quality between two tasks.
$\omega_1$ measures the feature adaptation quality of features originally extracted in the first task (\ie $y \in \cC^{1}$), showing how much the adapted features can diverge from their optimal representation due to accumulated error.
Adaptation quality is computed for all $L=500$ feature descriptors per class.

Figure~\ref{fig:task_cifar_quality} shows the adaptation quality $\omega_{t-1}$ and $\omega_1$ on CIFAR-100 with $M=10$.
We report the quality measures for when $P$ number images are also preserved in the memory.
We observe that $P=0$ achieves $\omega_{t-1}$ greater than $0.9$ in all tasks.
It increases as more classes are seen, most likely due to the network becoming more stable.
After $10$ tasks, $\omega_{1}$ is still close to $0.8$, indicating our feature adaptation is still relatively successful after training $9$ subsequent tasks with no preserved images.
Adaptation quality improves as $P$ increases, showing that preserving images also helps with learning a better feature adaptation.

\section{Balanced Feature Classifier Training}
\label{sec:balance}
Wu~\etal~\cite{WCW+19} illustrated that training a classifier on fewer examples for previous classes introduces a bias towards new classes. To verify the robustness of our method under this setting, we investigate the effect of training our feature classifier on class-balanced and class-imbalanced training sets.

In our main experiments, we train our feature classifier $g\prm{\tilde{W}}$ with a balanced number of examples per class. We repeat our ImageNet-100 experiment in Table \ref{tab:sota_table} without balancing the classifier training samples. In the unbalanced setting, the old classes contain $250$ features per class in memory, whereas the new classes each contain $\sim 1300$ feature vectors. In the balanced setting, all classes contain $250$ feature vectors. On ImageNet-100, we achieve 0.893 accuracy with class-imbalanced training, compared to 0.913 accuracy with class-balanced training (reported in Table \ref{tab:sota_table}). This shows that even though more training data is utilized in the class-imbalanced setting, the imbalanced class bias leads to a drop in overall performance.

Our method addresses this problem by building a large balanced feature set for training. Our feature adaptation method not only reduces the memory footprint compared to \cite{WCW+19} (see Table \ref{tab:sota_table}), but also allows substantially more stored data points from old classes (250 features per class compared to 20 images per class for \cite{WCW+19}). This may explain some improvement in our results over previous methods. Lastly, the significant increase in number of stored examples provides flexibility to remove examples to keep classes balanced.

\end{document}